\documentclass[11pt,a4paper]{article}
\usepackage[hyperref]{emnlp2018}
\usepackage{times}
\usepackage{latexsym}
\usepackage{graphicx}
\usepackage{amsmath}
\usepackage{amssymb}
\usepackage{color}
\usepackage{booktabs}
\usepackage{xspace}
\usepackage{enumitem}
\usepackage{multirow}
\usepackage{makecell}
\usepackage[normalem]{ulem}
\usepackage{soul}
\usepackage{cancel}

\usepackage{url}
\usepackage{graphicx} 
\usepackage[caption=false]{subfig}
\graphicspath{ {images/} }

\aclfinalcopy

\usepackage{amsmath}
\usepackage{amsfonts}
\newcommand{\R}{\mathbb{R}}
\newcommand{\softmax}{\text{softmax}}
\newcommand{\decCell}{\text{LSTM}}

\def\figref#1{Fig.~\ref{#1}}
\def\secref#1{Sec.~\ref{#1}}
\def\tabref#1{Table~\ref{#1}} 

\newcommand{\baseline}{Multi-Hop Pointer-Generator Model}
\newcommand{\baselineAbbv}{MHPGM}
\newcommand{\fullModel}{NOIC}
\newcommand{\fullModelFullName}{Necessary and Optional Information Cell~}
\newcommand{\prob}{\mathbb{P}}
\newcommand\bvect[1]{{\bf #1}}

\newcommand\blfootnote[1]{%
  \begingroup
  \renewcommand\thefootnote{}\footnote{#1}%
  \addtocounter{footnote}{-1}%
  \endgroup
}

\title{Commonsense for Generative Multi-Hop Question Answering Tasks}

\author{Lisa Bauer$^*$ \;\;\;\;\;\;\; Yicheng Wang$^*$ \;\;\;\;\;\;\; Mohit Bansal \\
  UNC Chapel Hill \\
  {\tt \{lbauer6, yicheng, mbansal\}@cs.unc.edu} \\
 }
 
\date{}
 
\begin{document}
\maketitle
\begin{abstract}

Reading comprehension QA tasks have seen a recent surge in popularity, yet most works have focused on fact-finding extractive QA. We instead focus on a more challenging multi-hop generative task (NarrativeQA), which requires the model to reason, gather, and synthesize disjoint pieces of information within the context to generate an answer.
This type of multi-step reasoning also often requires understanding implicit relations, which humans resolve via  external, background commonsense knowledge.
We first present a strong generative baseline that uses a multi-attention mechanism to perform multiple hops of reasoning and a pointer-generator decoder to synthesize the answer. This model performs substantially better than previous generative models, and is competitive with current state-of-the-art span prediction models. We next introduce a novel system for selecting grounded multi-hop relational commonsense information from ConceptNet via a pointwise mutual information and term-frequency based scoring function.
Finally, we effectively use this extracted commonsense information to fill in gaps of reasoning between context hops, using a selectively-gated attention mechanism.
This boosts the model's performance significantly (also verified via human evaluation), establishing a new state-of-the-art for the task. 
We also show promising initial results of the generalizability of our background knowledge enhancements by demonstrating some improvement on QAngaroo-WikiHop, another multi-hop reasoning dataset.

\end{abstract}

\section{Introduction}
In this paper, we explore the task of machine reading comprehension (MRC) based QA.\blfootnote{$*$ Equal contribution (published at EMNLP 2018).}\blfootnote{We publicly release all our code, models, and data at: \scriptsize{\url{https://github.com/yicheng-w/CommonSenseMultiHopQA}}}
This task tests a model's natural language understanding capabilities by asking it to answer a question based on a passage of relevant content.
Much progress has been made in reasoning-based MRC-QA on the bAbI dataset~\cite{Babi},
which contains questions that require
the combination of multiple disjoint pieces of evidence in the context.
However, due to its synthetic nature, bAbI evidences have smaller lexicons and simpler passage structures when compared to human-generated text.

There also have been several attempts at the MRC-QA task on human-generated text. Large scale datasets such as CNN/DM~\cite{cnndm} and SQuAD~\cite{rajpurkar2016squad} have made the training of end-to-end neural models possible.
However, these datasets are fact-based and do not place heavy emphasis on multi-hop reasoning capabilities.
More recent datasets such as QAngaroo~\cite{welbl2017constructing} have prompted a strong focus on multi-hop reasoning in very long texts.
However, QAngaroo is an extractive dataset where answers are guaranteed to be spans within the context; hence, this is more focused on fact finding and linking, and does not require models to synthesize and generate new information.

We focus on the recently published NarrativeQA generative dataset~\cite{kovcisky2017narrativeqa} that contains questions requiring multi-hop reasoning for long, complex stories and other narratives, which requires the model to go beyond fact linking and to synthesize non-span answers.
Hence, models that perform well on previous reasoning tasks~\cite{dhingra2018neural} have had limited success on this dataset.
In this paper, we first propose the \baseline\ (\baselineAbbv), a strong baseline model that uses multiple hops of bidirectional attention, self-attention, and a pointer-generator decoder to effectively read and reason within a long passage and synthesize a coherent response.
Our model achieves 41.49 Rouge-L and 17.33 METEOR on the summary subtask of NarrativeQA, substantially better than the performance of previous generative models.

Next, to address the issue that understanding human-generated text and performing long-distance reasoning on it often involves intermittent access to missing hops of external commonsense (background) knowledge, we present an algorithm for selecting useful, grounded multi-hop relational knowledge paths from ConceptNet~\cite{speer2012representing} via a pointwise mutual information (PMI) and term-frequency-based scoring function.
We then present a novel method of inserting these selected commonsense paths between the hops of document-context reasoning within our model, via the \fullModelFullName (\fullModel), which employs a selectively-gated attention mechanism that utilizes commonsense information to effectively fill in gaps of inference.
With these additions, we further improve performance on the NarrativeQA dataset, achieving 44.16 Rouge-L and 19.03 METEOR (also verified via human evaluation). We also provide manual analysis on the effectiveness of our commonsense selection algorithm.

Finally, to show the generalizability of our multi-hop reasoning and commonsense methods, we show some promising initial results via the addition of commonsense information over the baseline on QAngaroo-WikiHop~\cite{welbl2017constructing}, an extractive dataset for multi-hop reasoning from a different domain.

\section{Related Work}
\label{sec:related}

\noindent\textbf{Machine Reading Comprehension:}
MRC has long been a task used to assess a model's ability to understand and reason about language.
Large scale datasets such as CNN/Daily Mail ~\cite{cnndm} and SQuAD~\cite{rajpurkar2016squad} have encouraged the development of many advanced, high performing attention-based neural models~\cite{seo2016bidirectional,dhingra2016gated}.
Concurrently, datasets such as bAbI~\cite{Babi} have focused specifically on multi-step reasoning by requiring the model to reason with disjoint pieces of information.
On this task, it has been shown that iteratively updating the query representation with information from the context can effectively emulate multi-step reasoning~\cite{sukhbaatar2015end}.

More recently, there has been an increase in multi-paragraph, multi-hop inference QA datasets such as QAngaroo~\cite{welbl2017constructing} and NarrativeQA~\cite{kovcisky2017narrativeqa}.
These datasets have much longer contexts than previous datasets, and answering a question often requires the synthesis of multiple discontiguous pieces of evidence.
It has been shown that models designed for previous tasks~\cite{seo2016bidirectional,kadlec2016text} have limited success on these new datasets.
In our work, we expand upon Gated Attention Network~\cite{dhingra2016gated} to create a baseline model better suited for complex MRC datasets such as NarrativeQA by improving its attention and gating mechanisms, expanding its generation capabilities, and allowing access to external commonsense for connecting implicit relations.

\noindent\textbf{Commonsense/Background Knowledge:}
Commonsense or background knowledge has been used for several tasks including opinion mining~\cite{cambria2010sentic}, sentiment analysis \cite{poria2015sentiment,poria2016sentic}, handwritten text recognition \cite{wang2013common}, and more recently, dialogue~\cite{young2017augmenting,ghazvininejad2017knowledge}.
These approaches add commonsense knowledge as relation triples or features from external databases. 
Recently, large-scale graphical commonsense databases such as ConceptNet~\cite{speer2012representing} use graphical structure to express intricate relations between concepts, but effective goal-oriented graph traversal has not been extensively used in previous commonsense incorporation efforts.
Knowledge-base QA is a task in which systems are asked to find answers to questions by traversing knowledge graphs~\cite{bollacker2008freebase}. Knowledge path extraction has been shown to be effective at the task~\cite{bordes2014question,constraint_bao}.
We apply these techniques to MRC-QA by using them to extract useful commonsense knowledge paths that fully utilize the graphical nature of databases such as ConceptNet~\cite{speer2012representing}.

\noindent\textbf{Incorporation of External Knowledge:}
There have been several attempts at using external knowledge to boost model performance on a variety of tasks:
\newcite{chen2018natural} showed that adding lexical information from semantic databases such as WordNet improves performance on NLI; \newcite{xu2016incorporating} used a gated recall-LSTM mechanism to incorporate commonsense information into token representations in dialogue.

In MRC, \newcite{weissenborn2017dynamic} integrated external background knowledge into an NLU model by using contextually-refined word embeddings which integrated information from ConceptNet (single-hop relations mapped to unstructured text) via a single layer bidirectional LSTM.
Concurrently to our work,~\newcite{mihaylov2018knowledgeable} showed improvements on a cloze-style task by incorporating commonsense knowledge via a context-to-commonsense attention, where commonsense relations were extracted as triples. This work represented commonsense relations as key-value pairs and combined context representation and commonsense via a static gate.

Differing from previous works, we employ multi-hop commonsense paths (multiple connected edges within ConceptNet graph that give us information beyond a single relationship triple) to help with our MRC model. Moreover, we use this in tandem with our multi-hop reasoning architecture to incorporate different aspects of the commonsense relationship path at each hop, in order to bridge different inference gaps in the multi-hop QA task. Additionally, our model performs synthesis with its external, background knowledge as it generates, rather than extracts, its answer.

\begin{figure*}[t] 
	\centering
    \includegraphics[clip,width=0.95\textwidth]{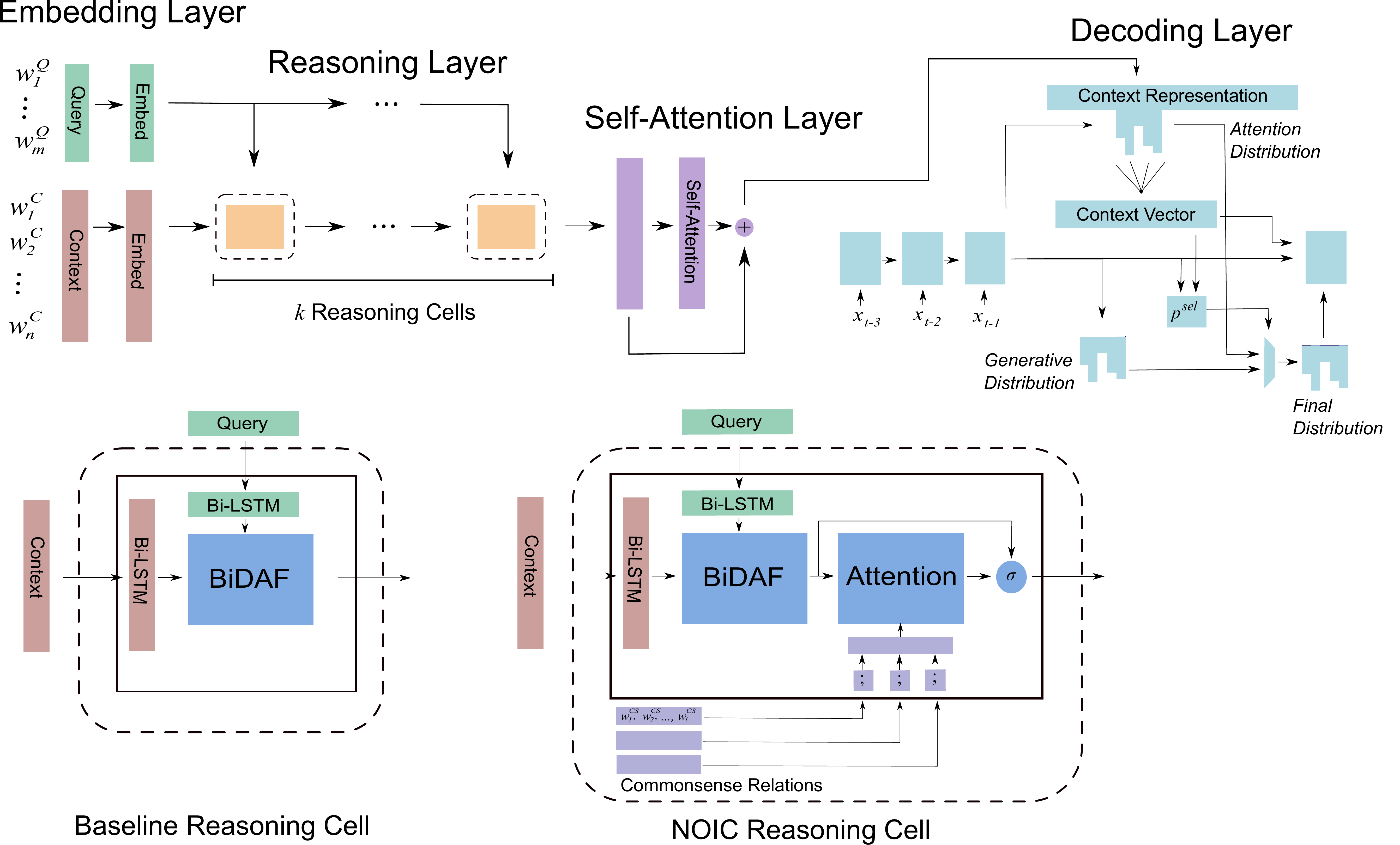}
    \vspace{-10pt}
    \caption{Architecture for our \baseline, and the \fullModel\ commonsense reasoning cell.
        \label{fig:model_figure}
        \vspace{-5pt}
    }
\end{figure*}

\section{Methods}
\label{sec:methods}

\subsection{Multi-Hop Pointer-Generator Baseline}
\label{sec:models}

We first rigorously state the problem of generative QA as follows: given two sequences of input tokens: the context, $X^C = \{w_1^C, w_2^C, \dots, w_n^C\}$ and the query, $X^Q = \{w_1^Q, w_2^Q, \dots, w_m^Q\}$, the system should generate a series of answer tokens $X^a = \{w_1^a, w_2^a, \dots, w_p^a\}$.
As outlined in previous sections, an effective generative QA model needs to be able to perform several hops of reasoning over long and complex passages.
It would also need to be able to generate coherent statements to answer complex questions while having the ability to copy rare words such as specific entities from the reading context.
With these in mind, we propose the \baseline\ (\baselineAbbv) baseline, a novel combination of previous works with the following major components:
\begin{itemize}[leftmargin=*]
    \setlength\itemsep{-0.45em}
    \item \textbf{Embedding Layer}: The tokens are embedded into both learned word embeddings and pretrained context-aware embeddings (ELMo ~\cite{peters2018deep}).
    \item \textbf{Reasoning Layer}: The embedded context is then passed through $k$ reasoning cells, each of which iteratively updates the context representation with information from the query via BiDAF attention~\cite{seo2016bidirectional}, emulating a single reasoning step within the multi-step reasoning process.
    \item \textbf{Self-Attention Layer}: The context representation is passed through a layer of self-attention \cite{cheng2016long} to resolve long-term dependencies and co-reference within the context.
    \item \textbf{Pointer-Generator Decoding Layer}: A attention-pointer-generator decoder \cite{see2017get} that attends on and potentially copies from the context is used to create the answer.
\end{itemize}

The overall model is illustrated in \figref{fig:model_figure}, and the layers are described in further detail below.

\noindent\textbf{Embedding layer}:
We embed each word from the context and question with a learned embedding space of dimension $d$.
We also obtain context-aware embeddings for each word via the pre-trained embedding from language models (ELMo) (1024 dimensions).
The embedded representation for each word in the context or question, $\bvect{e}^C_i$ or $\bvect{e}^Q_i \in \R^{d + 1024}$, is the concatenation of its learned word embedding and ELMo embedding.

\noindent\textbf{Reasoning layer}: Our reasoning layer is composed of $k$ reasoning cells (see \figref{fig:model_figure}), where each incrementally updates the context representation.
The $t$\textsuperscript{th} reasoning cell's inputs are the previous step's output ($\{\bvect{c}^{t-1}_i\}_{i=1}^n$) and the embedded question ($\{\bvect{e}^Q_i\}_{i=1}^m$). It first creates step-specific context and query encodings via cell-specific bidirectional LSTMs:
\vspace{-3pt}
\[\bvect{u}^t = \text{BiLSTM}(\bvect{c}^{t-1}); \qquad \bvect{v}^t = \text{BiLSTM}(\bvect{e}^Q)\]
Then, we use bidirectional attention~\cite{seo2016bidirectional} to emulate a hop of reasoning by focusing on relevant aspects of the context.
Specifically, we first compute context-to-query attention:
\[S^t_{ij} = W_1^t \bvect{u}_i^t + W_2^t \bvect{v}_j^t + W_3^t (\bvect{u}_i^t \odot \bvect{v}_j^t)\]
\[p^t_{ij} = \frac{\exp(S_{ij}^t)}{\sum_{k=1}^m \exp(S_{ik}^t)}\]
\[(\bvect{c_q})_i^t = \sum_{j=1}^m p^t_{ij} \bvect{v}_j^t\]
where $W_1^t$, $W_2^t$, $W_3^t$ are trainable parameters, and $\odot$ is
elementwise multiplication.
We then compute a query-to-context attention vector:
\[m_i^t = \max_{1 \leq j \leq m} S^t_{ij}\]
\[p^t_i = \frac{\exp(m_i^t)}{\sum_{j=1}^n \exp(m_j^t)}\]
\[\bvect{q_c}^t = \sum_{i=1}^n p^t_i \bvect{u}_i^t\]
We then obtain the updated context representation:
\[\bvect{c}_i^{t} = [\bvect{u}_i^t ; (\bvect{c_q})_i^t ; \bvect{u}_i^t \odot (\bvect{c_q})_i^t ; \bvect{q_c}^t \odot (\bvect{c_q})_i^t]\]
where $;$ is concatenation, $\bvect{c}^t$ is the cell's output.

The initial input of the reasoning layer is the embedded context representation, i.e., $\bvect{c}^0 = \bvect{e}^C$, and the final output of the reasoning layer is the output of the last cell, $\bvect{c}^k$.

\noindent\textbf{Self-Attention Layer}:
As the final layer before answer generation, we utilize a residual static self-attention mechanism \cite{clark2017simple} to help the model process long contexts with long-term dependencies.
The input of this layer is the output of the last reasoning cell, $\bvect{c}^k$. We first pass this representation through a fully-connected layer and then a bi-directional LSTM to obtain another representation of the context $\bvect{c}^{SA}$.
We obtain the self attention representation $\bvect{c'}$:
\[S_{ij}^{SA} = W_4 \bvect{c}^{SA}_i + W_5 \bvect{c}^{SA}_j + W_6 (\bvect{c}^{SA}_i \odot \bvect{c}^{SA}_j)\]
\[p_{ij}^{SA} = \frac{\exp(S_{ij}^{SA})}{\sum_{k=1}^n \exp(S_{ik}^{SA})}\]
\[\bvect{c'}_i = \sum_{j=1}^n p_{ij}^{SA} \bvect{c}^{SA}_j\]
where $W_4$, $W_5$, and $W_6$ are trainable parameters. 

The output of the self-attention layer is generated by another layer of bidirectional LSTM. 
\[\bvect{c''} = \text{BiLSTM}([\bvect{c'} ; \bvect{c}^{SA} ; \bvect{c'} \odot \bvect{c}^{SA}]\]
Finally, we add this residually to $\bvect{c}^k$ to obtain the encoded context $\bvect{c} = \bvect{c}^k + \bvect{c''}$.

\noindent\textbf{Pointer-Generator Decoding Layer}: 
Similar to the work of~\newcite{see2017get}, we use a pointer-generator model attending on (and potentially copying from) the context.

At decoding step $t$, 
the decoder receives the input $\bvect{x}_t$ (embedded representation of last timestep's output), the last time step's hidden state $\bvect{s}_{t-1}$ and context vector $\bvect{a}_{t-1}$.
The decoder computes the current hidden state $\bvect{s}_t$ as:
\[\bvect{s}_t = \decCell([\bvect{x}_t; \bvect{a}_{t-1}], \bvect{s}_{t-1})\]
This hidden state is then used to compute a probability distribution over the generative vocabulary:
\[P_{gen} = \softmax(W_{gen} \bvect{s}_t + \bvect{b}_{gen})\]

We employ Bahdanau attention mechanism~\cite{bahdanau2014neural} to attend over the context ($\bvect{c}$ being the output of self-attention layer):
\[\alpha_i = \bvect{v}^\intercal \tanh(W_c \bvect{c}_i + W_s \bvect{s}_t + b_{attn})\]
\[\hat \alpha_i = \frac{\exp(\alpha_i)}{\sum_{j=1}^n \exp(\alpha_j)}\]
\[\bvect{a}_t = \sum_{i=1}^n \hat \alpha_i \bvect{c}_i\]

We utilize a pointer mechanism that allows the decoder to directly copy tokens from the context based on $\hat \alpha_i$.
We calculate a selection distribution $\bvect{p}^{sel} \in \R^2$, where $p^{sel}_1$ is the probability of generating a token from $P_{gen}$ and $p^{sel}_2$ is the probability of copying a word from the context:
\[\bvect{o} = \sigma(W_a \bvect{a}_t + W_x \bvect{x}_t + W_s \bvect{s}_t + b_{ptr})\]
\[\bvect{p}^{sel} = \softmax(\bvect{o})\]

Our final output distribution at timestep $t$ is a weighted sum of the generative distribution and the copy distribution:
\vspace{-3pt}
$$P_t(w) = p^{sel}_1 P_{gen}(w) + p^{sel}_2 \sum_{i: w^C_i = w} \hat \alpha_i $$
\vspace{-15pt}
\begin{figure}[t]
\includegraphics[width=8cm]{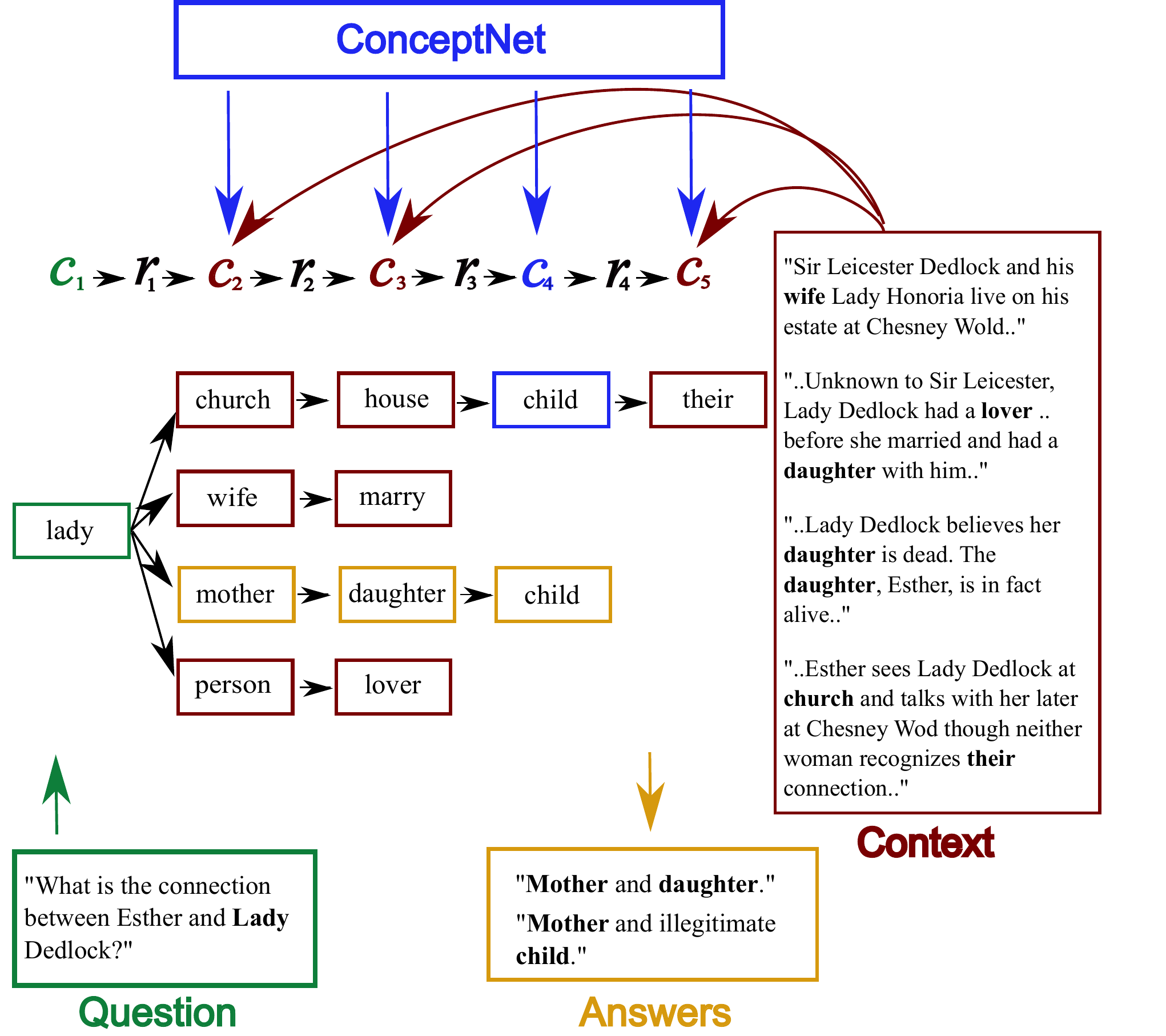}
\caption{Commonsense selection approach.
}
\label{sampling_method}
\vspace{-10pt}
\end{figure} 
\subsection{Commonsense Selection and Representation}
\label{sec:commonsense}

In QA tasks that require multiple hops of reasoning, the model often needs knowledge of relations not directly stated in the context to reach the correct conclusion.
In the datasets we consider, manual analysis shows that external knowledge is frequently needed for inference (see \tabref{tab:commonsense_req}).

Even with a large amount of training data, it is very unlikely that a model is able to learn every nuanced relation between concepts and apply the correct ones (as in~\figref{sampling_method}) when reasoning about a question.
We remedy this issue by introducing {\it grounded} commonsense (background) information using relations between concepts from ConceptNet~\cite{speer2012representing}\footnote
{A semantic network where the nodes are individual concepts (words or phrases) and the edges describe directed relations between them (e.g., $\langle \text{island, UsedFor, vacation} \rangle$).} that help inference by introducing useful connections between concepts in the context and question.

\begin{table}[t]
    \centering
    \begin{small}
    \begin{tabular}{cc}\toprule
    \textbf{Dataset} & \textbf{Outside Knowledge Required} \\
    \midrule
	WikiHop & 11\% \\
    NarrativeQA & 42\% \\
    \bottomrule
    \end{tabular}
    \end{small}
    \caption{Qualitative analysis of commonsense requirements. WikiHop results are from~\newcite{welbl2017constructing}; NarrativeQA results are from our manual analysis (on the validation set).}
    \label{tab:commonsense_req}
    \vspace{-10pt}
\end{table}

Due to the size of the semantic network and the large amount of unnecessary information, we need an effective way of selecting relations which provides novel information while being grounded by the context-query pair. Our commonsense selection strategy is twofold: (1) collect potentially relevant concepts via a tree construction method aimed at selecting with high recall candidate reasoning paths, and (2) rank and filter these paths to ensure both the quality and variety of added information via a 3-step scoring strategy (initial node scoring, cumulative node scoring, and path selection).  We will refer to \figref{sampling_method} as a running example throughout this section.\footnote{We release all our commonsense extraction code and the extracted commonsense data at: \scriptsize{\url{https://github.com/yicheng-w/CommonSenseMultiHopQA}}}

\subsubsection{Tree Construction}
Given context $C$ and question $Q$,
we want to construct paths grounded in the pair that emulate reasoning steps required to answer the question.
In this section, we build `prototype' paths by constructing trees rooted in concepts in the query with the following branching steps\footnote{If we are unable to find a relation that satisfies the condition, we keep the steps up to and including the node.} to emulate multi-hop reasoning process. For each concept $c_1$ in the question, we do: 

\noindent\textbf{Direct Interaction}:
In the first level, we select relations $r_1$ from ConceptNet that directly link $c_1$ to a concept within the context, $c_2 \in C$, e.g., in \figref{sampling_method}, we have {\it lady} $\rightarrow$ {\it church}, {\it lady} $\rightarrow$ {\it mother}, {\it lady} $\rightarrow$ {\it person}.

\noindent\textbf{Multi-Hop}:
We then select relations in ConceptNet $r_2$ that link $c_2$ to another concept in the context, $c_3 \in C$. This emulates a potential reasoning hop within the context of the MRC task, e.g., {\it church} $\rightarrow$ {\it house}, {\it mother}  $\rightarrow$ {\it daughter}, {\it person}  $\rightarrow$ {\it lover}.
    
\noindent\textbf{Outside Knowledge}:
We then allow an unconstrained hop into $c_3$'s neighbors in ConceptNet, getting to $c_4 \in \text{nbh}(c_3)$ via $r_3$ ($\text{nbh}(v)$ is the set of nodes that can be reached from $v$ in one hop). This emulates the gathering of useful external information to complete paths within the context, e.g., {\it house} $\rightarrow$ {\it child}, {\it daughter} $\rightarrow$ {\it child}.
    
\noindent\textbf{Context-Grounding}:
To ensure that the external knowledge is indeed helpful to the task, and also to explicitly link 2nd degree neighbor concepts within the context, we finish the process by grounding it again into context by connecting $c_4$ to $c_5 \in C$ via $r_4$, e.g., {\it child} $\rightarrow$ {\it their}.

\subsubsection{Rank and Filter}
This tree building process collects a large number of potentially relevant and useful paths. However, this step also introduces a large amount of noise. For example, given the question and full context (not depicted in the figure) in \figref{sampling_method}, we obtain the path ``{\it between} $\rightarrow$  {\it hard} $\rightarrow$  {\it being} $\rightarrow$  {\it cottage} $\rightarrow$  {\it country}''  using our tree building method, which is not relevant to our question.
Therefore, to improve the precision of useful concepts, we rank these knowledge paths by their relevance and filter out noise using the following 3-step scoring method:

\noindent\textbf{Initial Node Scoring}:
We want to select paths with nodes that are important to the context, in order to provide the most useful commonsense relations.
We approximate importance and saliency for concepts in the context by their term-frequency, under the heuristic that important concepts occur more frequently. Thus we score $c \in \{c_2, c_3, c_5\}$ by:
$\text{score}(c) = \text{count}(c) / |C|$,
where $|C|$ is the context length and $\text{count}()$ is the number of times a concept appears in the context. In \figref{sampling_method}, this ensures that concepts like {\it daughter} are scored highly due to their frequency in the context.

For $c_4$, we use a special scoring function as it is an unconstrained hop into ConceptNet.
We want $c_4$ to be a logically consistent next step in reasoning following the path of $c_1$ to $c_3$, e.g., in \figref{sampling_method}, we see that {\it child} is a logically consistent next step after the partial path of {\it mother} $\rightarrow$ {\it daughter}. We approximate this based on the heuristic that logically consistent paths occur more frequently.
Therefore, we score this node via Pointwise Mutual Information (PMI) between the partial path $c_{1-3}$ and $c_4$: $\text{PMI}(c_4, c_{1-3}) = \log(\prob(c_4, c_{1-3})/\prob(c_4) \prob(c_{1-3}))$, where
\vspace{-2pt}
\begin{align*}
\prob(c_4, c_{1-3}) &= \frac{\text{\# of paths connecting } c_1, c_2, c_3, c_4}{\text{\# of distinct paths of length 4}}\\
\prob(c_4) &= \frac{\text{\# of nodes that can reach } c_4}{|\text{ConceptNet}|} \\
\prob(c_{1-3}) &= \frac{\text{\# of paths connecting } c_1, c_2, c_3}{\text{\# of paths of length 3}}
\end{align*}

Further, it is well known that PMI has high sensitivity to low-frequency values, thus we use normalized PMI (NPMI) \cite{bouma2009normalized}:
$\text{score}(c_4) 
				  = \text{PMI}(c_4, c_{1-3}) / (-\log\prob(c_4, c_{1-3}))$.

Since the branching at each juncture represents a hop in the multi-hop reasoning process, and hops at different levels or with different parent nodes do not `compete' with each other, we normalize each node's score against its siblings:

\vspace{5pt}
$\text{n-score}(c) = \softmax_{\text{siblings}(c)}(\text{score}(c))$.
\vspace{5pt}

\noindent\textbf{Cumulative Node Scoring}:
We want to add commonsense paths consisting of multiple hops of relevant information, thus we re-score each node based not only on its relevance and saliency but also that of its tree descendants. 

We do this by computing a cumulative node score from the bottom up, where at the leaf nodes, we have $\text{c-score} = \text{n-score}$, and for $c_l$ not a leaf node, we have
$\text{c-score}(c_l) = \text{n-score}(c_l) + f(c_l)$
where $f$ of a node is the average of the c-scores of its top 2 highest scoring children.

For example, given the paths {\it lady} $\rightarrow$ {\it mother} $\rightarrow$ {\it daughter}, {\it lady} $\rightarrow$ {\it mother} $\rightarrow$ {\it married}, and {\it lady} $\rightarrow$ {\it mother} $\rightarrow$ {\it book}, we start the cumulative scoring at the leaf nodes, which in this case are {\it daughter}, {\it married}, and {\it book}, where  {\it daughter} and {\it married} are scored much higher than {\it book} due to their more frequent occurrences. Then, to cumulatively score ${\it mother}$, we would take the average score of its two highest scoring children (in this case {\it married} and {\it daughter})  and compound that with the score of {\it mother} itself. Note that the poor scoring of the irrelevant concept {\it book} does not affect the scoring of {\it mother}, which is quite high due to the concept's frequent occurrence and the relevance of its top scoring children.

\noindent\textbf{Path Selection}:
We select paths in a top-down breath-first fashion in order to add information relevant to different parts of the context.
Starting at the root, we recursively take two of its children with the highest cumulative scores until we reach a leaf, selecting up to $2^4=16$ paths. For example, if we were at node {\it mother}, this allows us to select the child node {\it daughter} and {\it married} over the child node {\it book}.
These selected paths, as well as their partial sub-paths, are what we add as external information to the QA model, i.e., we add the complete path $\langle$lady, AtLocation, church, RelatedTo, house, RelatedTo, child, RelatedTo, their$\rangle$, but also truncated versions of the path, including $\langle$lady, AtLocation, church, RelatedTo, house, RelatedTo, child$\rangle$.
We directly give these paths to the model as sequences of tokens.\footnote{In cases where more than one relation can be used to make a hop, we pick one at random.}

Overall, our sampling strategy provides the knowledge that a {\it lady} can be a  {\it mother} and that {\it mother} is connected to {\it daughter}. This creates a logical connection between {\it lady} and {\it daughter} which helps highlight the importance of our second piece of evidence (see \figref{sampling_method}). Likewise, the commonsense information we extracted create a similar connection in our third piece of evidence, which states the explicit connection between {\it daughter} and {\it Esther}.
We also successfully extract a more story context-centric connection, in which commonsense provides the knowledge that a {\it lady} is at the location {\it church}, which directs to another piece of evidence in the context. Additionally, this path also encodes a relation between  {\it lady} and {\it child}, by way of {\it church}, which is how {\it lady} and {\it Esther} are explicitly connected in the story.

\subsection{Commonsense Model Incorporation}
\label{sec:commonsense_model_inc}

Given the list of commonsense logic paths as sequences of words: $X^{CS}$ $=$ $\{{w}^{CS}_1$, ${w}^{CS}_2$, $\dots$, ${w}^{CS}_l\}$ where ${w}^{CS}_i$ represents the list of tokens that make up a single path, we first embed these commonsense tokens into the learned embedding space used by the model, giving us the embedded commonsense tokens, $\bvect{e}^{CS}_{ij} \in \R^d$.
We want to use these commonsense paths to fill in the gaps of reasoning between hops of inference.
Thus, we propose \fullModelFullName\ (\fullModel), a variation of our base reasoning cell used in the reasoning layer that is capable of incorporating optional helpful information.
 
\begin{table*}[t]
    \centering
    \begin{small}
    \begin{tabular}{lccccc}
    \toprule
    \textbf{Model} & \textbf{BLEU-1} & \textbf{BLEU-4} & \textbf{METEOR} & \textbf{Rouge-L} & \textbf{CIDEr}\\
    \midrule
    Seq2Seq \cite{kovcisky2017narrativeqa} & 15.89 & 1.26 & 4.08 & 13.15 & - \\
    ASR \cite{kovcisky2017narrativeqa} & 23.20 & 6.39 & 7.77 & 22.26 & -\\
    BiDAF$^\dagger$ \cite{kovcisky2017narrativeqa} & 33.72 & 15.53 & 15.38 & 36.30 & -\\
    BiAttn + MRU-LSTM$^\dagger$ \cite{tay2018multi} & 36.55 & 19.79 & 17.87 & 41.44 & -\\
    \midrule\midrule
    \baselineAbbv & 40.24 & 17.40 & 17.33 & 41.49 & 139.23 \\
    \baselineAbbv + \fullModel & \bf{43.63} & \bf{21.07} & \bf{19.03} & \bf{44.16} & \bf{152.98} \\
    \bottomrule
    \end{tabular}
    \end{small}
    \vspace{-5pt}
    \caption{Results across different metrics on the test set of NarrativeQA-summaries task.
    $^\dagger$ indicates span prediction models trained on the Rouge-L retrieval oracle.}
    \label{tab:resultsNQA}
    \vspace{-7pt}
\end{table*}

\begin{table}[t]
	\centering
    \begin{small}
    \begin{tabular}{lcc} \toprule
    \textbf{Model} & \textbf{Dev} & \textbf{Test}\\\midrule
    BiDAF \cite{welbl2017constructing} & 42.1 & 42.9 \\
    Coref-GRU \cite{dhingra2018neural} & 56.0 & 59.3 \\
    \midrule
    \baselineAbbv & 56.2 & 57.5 \\
    \baselineAbbv + \fullModel\ & \bf 58.5 & 57.9 \\
    \bottomrule
    \end{tabular}
    \end{small}
    \vspace{-5pt}
    \caption{Results of our models on WikiHop dataset, measured in \% accuracy.}
    \label{tab:wikihop_results}
    \vspace{-8pt}
\end{table}

\paragraph{\fullModel}
This cell is an extension to the base reasoning cell that allows the model to use commonsense information to fill in gaps of reasoning.
An example of this is on the bottom left of~\figref{fig:model_figure}, where we see that the cell first performs the operations done in the base reasoning cell and then adds optional, commonsense information.

At reasoning step $t$, after obtaining the output of the base reasoning cell, $\bvect{c}^t$, we create a cell-specific representation for commonsense information
by concatenating the embedded commonsense paths so that each path has a single vector representation, $\bvect{u}^{CS}_i$.
We then project it to the same dimension as $\bvect{c}^t_i$:
\(\bvect{v}^{CS}_i = \text{ReLU}(W \bvect{u}_i^{CS} + b)\)
where $W$ and $b$ are trainable parameters.

We use an attention layer to model the interaction between commonsense and the context:
\[S_{ij}^{CS} = W_1^{CS} \bvect{c}^t_i + W_2^{CS} \bvect{v}^{CS}_j + W_3^{CS} (\bvect{c}^t_i \odot \bvect{v}^{CS}_j)\]
\[p_{ij}^{CS} = \frac{\exp(S^{CS}_{ij})}{\sum_{k=1}^l \exp(S^{CS}_{ij})}\]
\[\bvect{c}^{CS}_i = \sum_{j=1}^l p_{ij}^{CS} \bvect{v}^{CS}_j\]

Finally, we combine this commonsense-aware context representation with the original $\bvect{c}^t_i$ via a sigmoid gate, since commonsense information is often not necessary at every step of inference:
\[\bvect{z}_i = \sigma(W_z [\bvect{c}^{CS}_i; \bvect{c}^t_i] + b_z)\]
\[(\bvect{c_o})^t_i = \bvect{z}_i \odot \bvect{c}^t_i + (1 - \bvect{z}_i) \odot \bvect{c}^{CS}_i\]
We use $\bvect{c_o}^t$ as the output of the current reasoning step instead of $\bvect{c}^t$.
As we replace each base reasoning cell with \fullModel, we selectively incorporate commonsense at every step of inference.

\section{Experimental Setup}

\noindent\textbf{Datasets:} We report results on two multi-hop reasoning datasets: generative NarrativeQA~\cite{kovcisky2017narrativeqa} (summary subtask) and extractive QAngaroo WikiHop~\cite{welbl2017constructing}.
For multiple-choice WikiHop, we rank candidate responses by their generation
probability. Similar to previous works ~\cite{dhingra2018neural}, we use
the non-oracle, unmasked and not-validated dataset.

\noindent\textbf{Evaluation Metrics:}
We evaluate NarrativeQA on the metrics proposed by its original authors: Bleu-1, Bleu-4~\cite{papineni2002bleu}, METEOR~\cite{banerjee2005meteor} and Rouge-L~\cite{lin2004rouge}.
We also evaluate on CIDEr~\cite{vedantam2015cider} which emphasizes annotator consensus.
For WikiHop, we evaluate on accuracy.

More dataset, metric, and all other training details are in the supplementary.

\section{Results}
\label{sec:results}
\subsection{Main Experiment}

The results of our model on both NarrativeQA and WikiHop with and without commonsense incorporation are shown in \tabref{tab:resultsNQA} and \tabref{tab:wikihop_results}.
We see empirically that our model outperforms all generative models on NarrativeQA, and is competitive with the top span prediction models.
Furthermore, with the \fullModel\ commonsense integration, we were able to further improve performance ($p<0.001$ on all metrics\footnote{Stat. significance computed using bootstrap test with 100K iterations~\cite{noreen1989computer,efron1994introduction}.}), establishing a new state-of-the-art for the task.

We also see that our model performs reasonably well on WikiHop, and further achieves promising initial improvements via the addition of commonsense, hinting at the generalizability of our approaches. We speculate that the improvement is smaller on Wikihop because only approximately 11\% of WikiHop data points require commonsense and because WikiHop data requires more fact-based commonsense (e.g., from Freebase~\cite{bollacker2008freebase}) as opposed to semantics-based commonsense (e.g., from ConceptNet~\cite{speer2012representing}).\footnote{All results here are for the standard (non-oracle) unmasked and not-validated dataset. \newcite{welbl2017constructing} has reported higher numbers on different data settings which are not comparable to our results.}

\subsection{Model Ablations}

We also tested the effectiveness of each component of our architecture as well as the effectiveness of adding commonsense information on the NarrativeQA validation set, with results shown in \tabref{tab:model_ablations}.
Experiment 1 and 5 are our models presented in \tabref{tab:resultsNQA}.
Experiment 2 demonstrates the importance of multi-hop attention by showing that if we only allow one hop of attention (even with all other components of the model, including ELMo embeddings) the model's performance decreases by over 12 Rouge-L points.
Experiment 3 and 4 demonstrate the effectiveness of other parts of our model. We
see that ELMo embeddings~\cite{peters2018deep} were also important for the
model's performance
and that self-attention is able to
contribute significantly to performance on top of other components of the model.
Finally, we see that effectively introducing external knowledge via our commonsense selection algorithm and \fullModel\ can improve performance even further on top of our strong baseline.

\subsection{Commonsense Ablations}

We also conducted experiments testing the effectiveness of our commonsense
selection and incorporation techniques.  We first tried to naively add
ConceptNet information by initializing the word embeddings with the
ConceptNet-trained embeddings, NumberBatch~\cite{speer2012representing} (we also
change embedding size from 256 to 300).  Then, to verify the effectiveness of
our commonsense selection and grounding algorithm, we test our best model on
in-domain noise by giving each context-query pair a set of random relations
grounded in other context-query pairs. This should teach the model about
general commonsense relations present in the domain of NarrativeQA but does not
provide grounding that fills in specific hops of inference. We also
experimented with a simpler commonsense extraction method of using a single hop
from the query to the context. The results of
these are shown in \tabref{tab:commonsense_abl_exps},
where we see that neither NumberBatch nor random-relationships nor single-hop commonsense offer
statistically significant improvements\footnote{The improvement in Rouge-L and METEOR for all three ablation approaches have
$p \geq 0.15$ with the bootstrap test.},
whereas our commonsense selection and incorporation mechanism improves performance significantly
across all metrics.
We also present several examples of extracted commonsense and its model attention visualization in the supplementary.

\begin{table}
	\centering
    \begin{small}
    \begin{tabular}{ccccccc}\toprule
    \textbf{\#} & \textbf{Ablation} & \textbf{B-1} & \textbf{B-4} & \textbf{M} & \textbf{R} & \textbf{C}\\
    \midrule
    1 & - & 42.3 & 18.9 & 18.3 & 44.9 & 151.6 \\
    2 & $k=1$ & 32.5 & 11.7 & 12.9 & 32.4 & 95.7 \\
    3 & - ELMo & 32.8 & 12.7 & 13.6 &  33.7 & 103.1 \\
    4 & - Self-Attn & 37.0 & 16.4 & 15.6 & 38.6 & 125.6 \\
    5 & + \fullModel & \bf 46.0 & \bf 21.9 & \bf 20.7 & \bf 48.0 & \bf 166.6 \\
    \bottomrule
    \end{tabular}
    \end{small}
    \caption{Model ablations on NarrativeQA val-set.}
    \label{tab:model_ablations}
\end{table}

\begin{table}
	\centering
    \begin{small}
    \begin{tabular}{lccccc}\toprule
    \textbf{Commonsense} & \textbf{B-1} & \textbf{B-4} & \textbf{M} & \textbf{R} & \textbf{C} \\
    \midrule
    None & 42.3 & 18.9 & 18.3 & 44.9 & 151.6 \\
    NumberBatch & 42.6 & 19.6 & 18.6 & 44.4 & 148.1 \\
    Random Rel. & 43.3 & 19.3 & 18.6 & 45.2 & 151.2 \\
    Single Hop & 42.1 & 19.9 & 18.2 & 44.0 & 148.6 \\
    Grounded Rel. & \bf 45.9 & \bf 21.9 & \bf 20.7 & \bf 48.0 & \bf 166.6\\
    \bottomrule
    \end{tabular}
    \end{small}
    \vspace{-5pt}
    \caption{Commonsense ablations on NarrativeQA val-set.}
    \label{tab:commonsense_abl_exps}
    \vspace{-10pt}
\end{table}

\begin{table}[t]
    \centering
    \begin{small}
    \begin{tabular}{ccc}
    \toprule
    & \multicolumn{2}{c}{Commonsense Required} \\
    & Yes & No \\\midrule
	Relevant CS Extracted& 34\% & 14\%\\
    Irrelevant CS Extracted & 16\%  & 36\% \\
    \bottomrule
    \end{tabular}
    \end{small}
    \vspace{-5pt}
    \caption{NarrativeQA's commonsense requirements and effectiveness of commonsense selection algorithm.}
    \label{tab:commonsense_an}
\end{table}
\section{Human Evaluation Analysis}
\label{sec:discussion}
We also conduct human evaluation analysis on both the quality of the selected commonsense relations, as well as the performance of our final model.

\noindent\textbf{Commonsense Selection:} We conducted manual analysis on a 50 sample subset of the NarrativeQA test set to check the effectiveness of our commonsense selection algorithm.
Specifically, given a context-query pair, as well as the commonsense selected by our algorithm, we conduct two independent evaluations: (1) was any external commonsense knowledge necessary for answering the question?; (2) were the commonsense relations provided by our algorithm relevant to the question? 
The result for these two evaluations as well as how they overlap with each other are shown in \tabref{tab:commonsense_an}, where we see that 50\% of the cases required external commonsense knowledge, and on a majority (34\%) of those cases our algorithm was able to select the correct/relevant commonsense information to fill in gaps of inference.
We also see that in general, our algorithm was able to provide useful commonsense 48\% of the time.

\noindent\textbf{Model Performance:}
We also conduct human evaluation to verify that our commonsense incorporated model was indeed better than \baselineAbbv.
We randomly selected 100 examples from the NarrativeQA test set,
along with both models' predicted answers, and for each datapoint, we
asked 3 external human evaluators (fluent English speakers) to decide (without knowing
which model produced each response) if one is strictly better than the other, or
that they were similar in quality (both-good or both-bad).
As shown in~\tabref{tab:human_eval}, we see that the human evaluation
results are in agreement with that of the automatic evaluation metrics: our
commonsense incorporation has a reasonable impact on the overall
correctness of the model. The inter-annotator agreement had a Fleiss
$\kappa = 0.831$, indicating `almost-perfect' agreement between the
annotators~\cite{landis1977measurement}.

\begin{table}
	\centering
    \begin{small}
    \begin{tabular}{l|c}\toprule
        \baselineAbbv+\fullModel\ better & 23\% \\
        \baselineAbbv\ better & 15\% \\
        Indistinguishable (Both-good) & 41\% \\
        Indistinguishable (Both-bad) & 21\% \\
    \bottomrule
    \end{tabular}
    \end{small}
    \vspace{-5pt}
    \caption{Human evaluation on the output quality of the \baselineAbbv+\fullModel\ vs. \baselineAbbv\ in terms of correctness.}
    \label{tab:human_eval}
    \vspace{-8pt}
\end{table}

\section{Conclusion}
\label{sec:conclusion}
We present an effective reasoning-generative QA architecture that is a novel combination of previous work, which uses multiple hops of bidirectional attention and a pointer-generator decoder to effectively perform multi-hop reasoning and synthesize a coherent and correct answer.
Further, we introduce an algorithm to select grounded, useful paths of commonsense knowledge to fill in the gaps of inference required for QA, as well a Necessary and Optional Information Cell (NOIC) which successfully incorporates this information during multi-hop reasoning to achieve the new state-of-the-art on NarrativeQA.

\section*{Acknowledgments}
\vspace{-5pt}
We thank the reviewers for their helpful comments.
This work was supported by DARPA (YFA17-D17AP00022), Google Faculty Research Award, Bloomberg Data Science Research Grant, and NVidia GPU awards. The views contained in
this article are those of the authors and not of the funding agency.

\bibliography{emnlp}
\bibliographystyle{acl_natbib_nourl}

\appendix
\section{Supplemental Material}

\subsection{Experimental Setup}
\paragraph{Datasets}
We test our model with and without commonsense addition on two challenging datasets that require multi-hop reasoning and external knowledge: NarrativeQA~\cite{kovcisky2017narrativeqa} and QAngaroo-WikiHop~\cite{welbl2017constructing}. 
NarrativeQA is a generative QA dataset where the passages are either stories or summaries of stories, and the questions ask about complex aspects of the narratives such as event timelines, characters, relations between characters, etc.
Each question has two answers which are generated by human annotators and usually cannot be found in the passage directly.
We focus on the summary subtask in this paper, where summaries have lengths of up to 1000 words.

We also test our model on WikiHop, a fact based, multi-hop dataset.
Questions in WikiHop often require a model to read several documents in order to obtain an answer.
We focus on the multiple-choice part of WikiHop, where models are tasked with picking the correct response from a pool of candidates.
We rank candidate responses by calculating their generation probability based on our model.
As this is a multi-document QA task, we first rank the candidate documents via TF-IDF cosine distance with the question, and then take the top $k$ documents such that their combined length is less than 1300 words.

\paragraph{Evaluation Metrics}
We evaluate NarrativeQA on the metrics proposed by its original authors: Bleu-1, Bleu-4~\cite{papineni2002bleu}, METEOR~\cite{banerjee2005meteor} and Rouge-L~\cite{lin2004rouge}.
We also evaluate on CIDEr~\cite{vedantam2015cider} as it places emphasize on annotator consensus.
For WikiHop, we evaluate on accuracy.

\paragraph{Training Details}
In training for both datasets, we minimize the negative log probability of generating the ground-truth answer with the Adam optimizer~\cite{kingma2014adam} with an initial learning rate of 0.001, a dropout-rate of 0.2 (dropout is applied to the input of each RNN layer) and batch size of 24. We use 256 dimensional word embeddings and a hidden size of 128 for all RNNs and $k=3$ hops of multi attention. At inference time we use greedy decoding to generate the answer.
For both NarrativeQA and WikiHop, we reached these parameters via tuning on the full, official validation set.

\begin{table*}[!b]
\section*{\makecell[l]{Commonsense Extraction and Visualization Examples \\ \\ \\}} 
	\centering
    \begin{small}
  \begin{tabular}{|p{0.15\textwidth}|p{0.8\textwidth}|} 
    \hline
    Question & \makecell[l]{What shore does Michael's \textbf{corpse} wash \textbf{up} on? \\ } \\
    Context & \makecell[l]{\\"..as the play begins nora and cathleen receive word from the priest that a \\ \textbf{body}, that may be their brother michael, has washed up on shore in donegal, \\ the island farthest  \textbf{north} of their home island of inishmaan.."\\ \\} \\
    Answers & \makecell[l]{the shore of donegal / donegal} \\
    \makecell[l]{Extracted \\ Commonsense} & \makecell[l]{ \\
    up $\rightarrow$ RelatedTo $\rightarrow$ wind $\rightarrow$ Antonym $\rightarrow$ her $\rightarrow$ RelatedTo $\rightarrow$ person\\
up $\rightarrow$ RelatedTo $\rightarrow$ north $\rightarrow$ RelatedTo $\rightarrow$ up\\
wash $\rightarrow$ RelatedTo $\rightarrow$ up\\
up $\rightarrow$ Antonym $\rightarrow$ down\\
wash $\rightarrow$ RelatedTo $\rightarrow$ water $\rightarrow$ PartOf $\rightarrow$ sea $\rightarrow$ RelatedTo $\rightarrow$ fish\\
up $\rightarrow$ RelatedTo $\rightarrow$ wind\\
wash $\rightarrow$ RelatedTo $\rightarrow$ water $\rightarrow$ PartOf $\rightarrow$ sea\\
shore $\rightarrow$ RelatedTo $\rightarrow$ sea\\
wash $\rightarrow$ RelatedTo $\rightarrow$ body\\
wash $\rightarrow$ Antonym $\rightarrow$ making\\
up $\rightarrow$ Antonym $\rightarrow$ down $\rightarrow$ Antonym $\rightarrow$ up\\
wash $\rightarrow$ RelatedTo $\rightarrow$ water $\rightarrow$ PartOf $\rightarrow$ sea $\rightarrow$ MadeOf $\rightarrow$ water\\
up $\rightarrow$ RelatedTo $\rightarrow$ wind $\rightarrow$ Antonym $\rightarrow$ her\\
wash $\rightarrow$ RelatedTo $\rightarrow$ water\\
up $\rightarrow$ RelatedTo $\rightarrow$ south\\
shore $\rightarrow$ RelatedTo $\rightarrow$ sea $\rightarrow$ MadeOf $\rightarrow$ water $\rightarrow$ AtLocation $\rightarrow$ bucket $\rightarrow$ RelatedTo $\rightarrow$ horse\\
wash $\rightarrow$ RelatedTo $\rightarrow$ clothing\\
wash $\rightarrow$ RelatedTo $\rightarrow$ water $\rightarrow$ PartOf $\rightarrow$ sea $\rightarrow$ MadeOf $\rightarrow$ water $\rightarrow$ PartOf $\rightarrow$ sea\\
shore $\rightarrow$ RelatedTo $\rightarrow$ sea $\rightarrow$ MadeOf $\rightarrow$ water\\
wash $\rightarrow$ Antonym $\rightarrow$ getting\\
\textbf{up $\rightarrow$ RelatedTo $\rightarrow$ north}\\
\textbf{corpse $\rightarrow$ RelatedTo $\rightarrow$ body}\\
shore $\rightarrow$ RelatedTo $\rightarrow$ sea $\rightarrow$ MadeOf $\rightarrow$ water $\rightarrow$ AtLocation $\rightarrow$ fountain\\
corpse $\rightarrow$ RelatedTo $\rightarrow$ body $\rightarrow$ RelatedTo $\rightarrow$ corpse\\
corpse $\rightarrow$ RelatedTo $\rightarrow$ body $\rightarrow$ RelatedTo $\rightarrow$ water\\
wash $\rightarrow$ HasContext $\rightarrow$ west\\
up $\rightarrow$ RelatedTo $\rightarrow$ wind $\rightarrow$ Antonym $\rightarrow$ her $\rightarrow$ RelatedTo $\rightarrow$ person $\rightarrow$ MadeOf $\rightarrow$ water\\
up $\rightarrow$ RelatedTo $\rightarrow$ wind $\rightarrow$ AtLocation $\rightarrow$ sea\\
wash $\rightarrow$ RelatedTo $\rightarrow$ water $\rightarrow$ AtLocation $\rightarrow$ can\\
shore $\rightarrow$ RelatedTo $\rightarrow$ sea $\rightarrow$ MadeOf $\rightarrow$ water $\rightarrow$ AtLocation $\rightarrow$ bucket\\
wash $\rightarrow$ RelatedTo $\rightarrow$ will\\
shore $\rightarrow$ RelatedTo $\rightarrow$ sea $\rightarrow$ MadeOf $\rightarrow$ water $\rightarrow$ AtLocation $\rightarrow$ fountain $\rightarrow$ RelatedTo $\rightarrow$ water
    } \\
    \hline
  \end{tabular}
 \caption{Example 1 selected commonsense paths.}
 \label{tab:extractedex1}
\end{small}

\end{table*}

\begin{figure*}[h]
  \centering
  \includegraphics[width=0.9\textwidth,trim={0 11cm 0 0},clip]{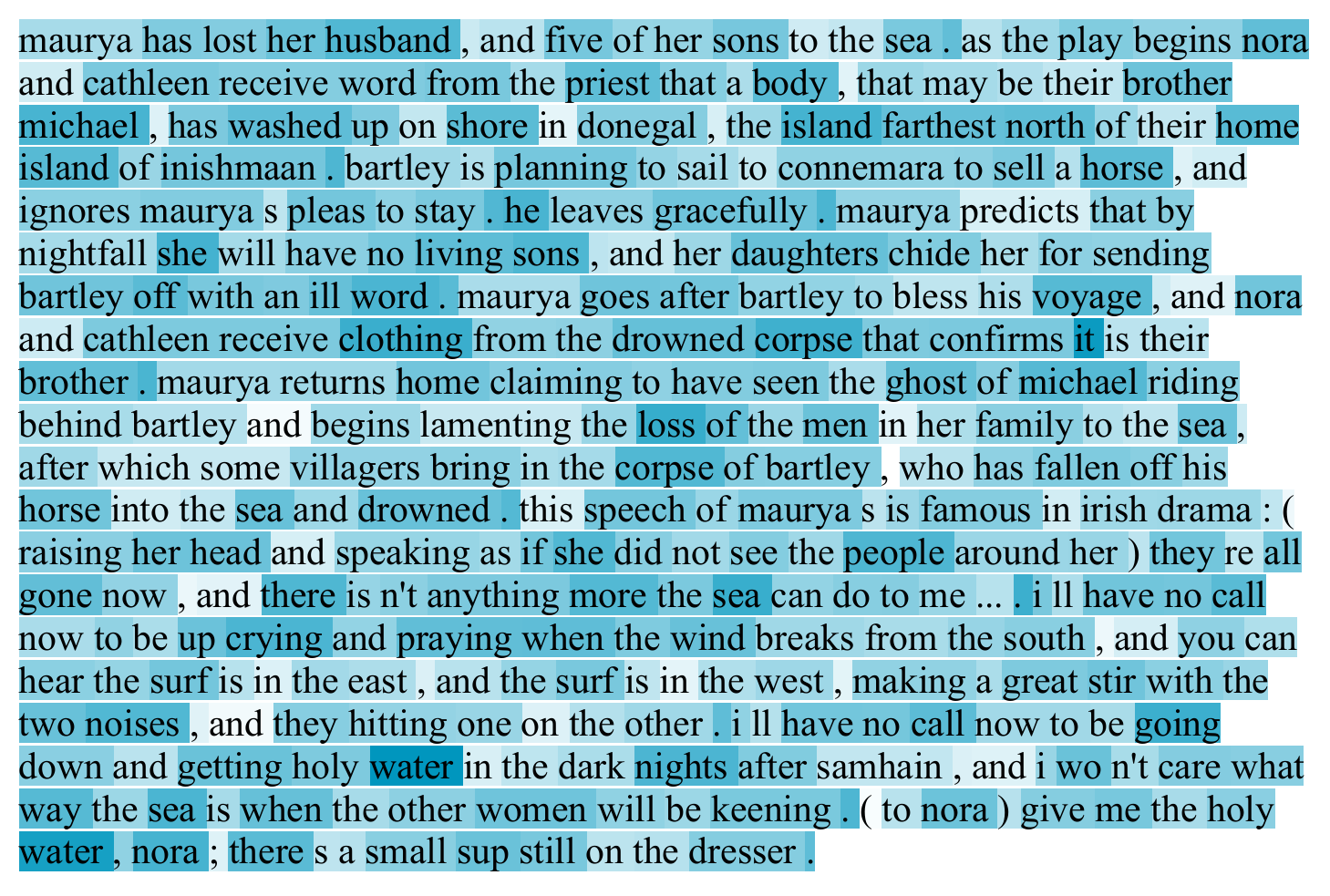}
 \caption{Example 1 visualized activation values of first attention hop ($1 - {\bf z}_1$).}
\end{figure*}
\begin{figure*}[h]
  \centering
  \includegraphics[width=0.9\textwidth,trim={0 11cm 0 0},clip]{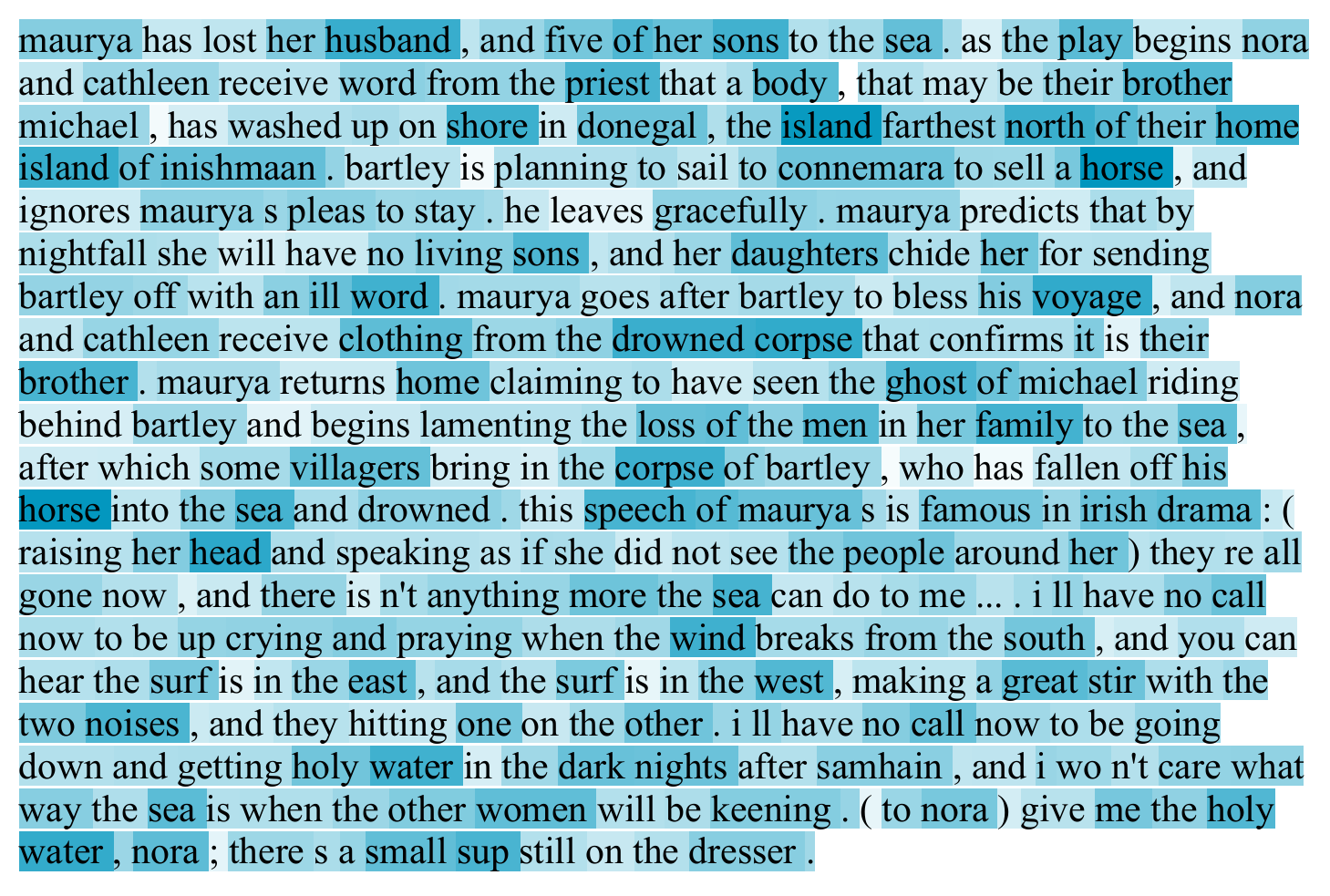}
 \caption{Example 1 visualized activation values of second attention hop ($1 - {\bf z}_2$).}
\end{figure*}
\begin{figure*}[h]
  \centering
  \includegraphics[width=0.9\textwidth,trim={0 11cm 0 0},clip]{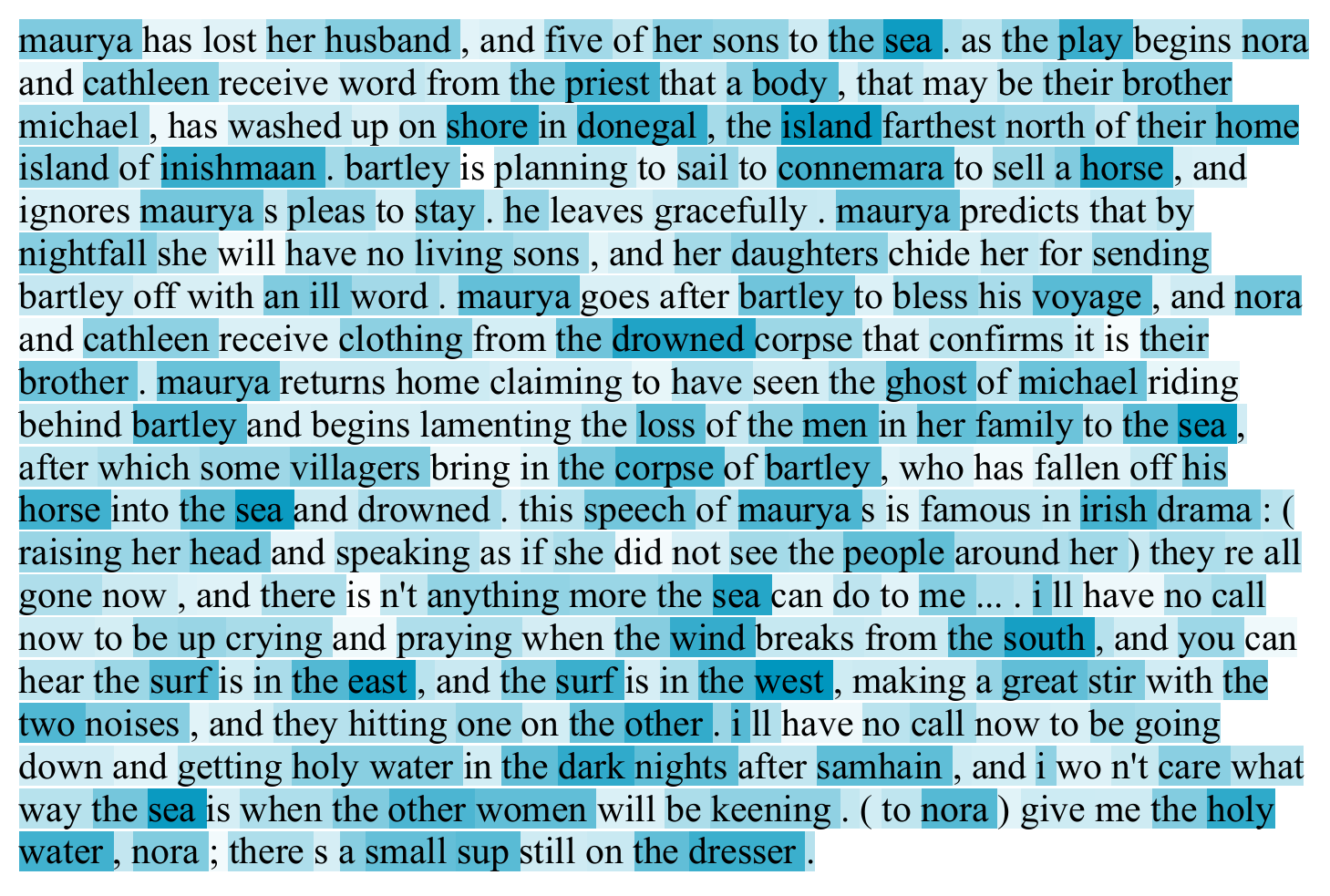}
 \caption{Example 1 visualized activation values of third attention hop ($1 - {\bf z}_3$).}
\end{figure*}
\begin{table*}[ht]
	\centering
    \begin{small}
  \begin{tabular}{|p{0.15\textwidth}|p{0.8\textwidth}|} 
    \hline
    Question & \makecell[l]{What \textbf{species} lives in the nearby mines?} \\
    Context & \makecell[l]{\\"..the nearby mines are inhabited by a \textbf{race} of goblins.."\\ \\} \\
    Answers & the goblins / goblins. \\
    \makecell[l]{Extracted \\ Commonsense} & \makecell[l]{\\
    species $\rightarrow$ RelatedTo $\rightarrow$ kingdom $\rightarrow$ RelatedTo $\rightarrow$ queen\\
species $\rightarrow$ RelatedTo $\rightarrow$ kingdom $\rightarrow$ RelatedTo $\rightarrow$ queen $\rightarrow$ UsedFor $\rightarrow$ people $\rightarrow$ HasA $\rightarrow$ feet\\
mines $\rightarrow$ FormOf $\rightarrow$ mine\\
lives $\rightarrow$ FormOf $\rightarrow$ life\\
mines $\rightarrow$ FormOf $\rightarrow$ mine $\rightarrow$ AtLocation $\rightarrow$ home $\rightarrow$ RelatedTo $\rightarrow$ person\\
species $\rightarrow$ RelatedTo $\rightarrow$ kingdom $\rightarrow$ RelatedTo $\rightarrow$ queen $\rightarrow$ UsedFor $\rightarrow$ people\\
species $\rightarrow$ RelatedTo $\rightarrow$ kingdom $\rightarrow$ DerivedFrom $\rightarrow$ king $\rightarrow$ RelatedTo $\rightarrow$ master\\
species $\rightarrow$ RelatedTo $\rightarrow$ kingdom $\rightarrow$ RelatedTo $\rightarrow$ queen $\rightarrow$ RelatedTo $\rightarrow$ person $\rightarrow$ Desires $\rightarrow$ feet\\
mines $\rightarrow$ FormOf $\rightarrow$ mine $\rightarrow$ AtLocation $\rightarrow$ home $\rightarrow$ RelatedTo $\rightarrow$ line $\rightarrow$ RelatedTo $\rightarrow$ thread\\
species $\rightarrow$ RelatedTo $\rightarrow$ kingdom $\rightarrow$ DerivedFrom $\rightarrow$ king $\rightarrow$ RelatedTo $\rightarrow$ leader $\rightarrow$ AtLocation \\ $\rightarrow$ company\\
species $\rightarrow$ RelatedTo $\rightarrow$ kingdom\\
species $\rightarrow$ RelatedTo $\rightarrow$ kingdom $\rightarrow$ DerivedFrom $\rightarrow$ king $\rightarrow$ RelatedTo $\rightarrow$ leader\\
species $\rightarrow$ RelatedTo $\rightarrow$ kingdom $\rightarrow$ DerivedFrom $\rightarrow$ king\\
mines $\rightarrow$ FormOf $\rightarrow$ mine $\rightarrow$ AtLocation $\rightarrow$ home $\rightarrow$ RelatedTo $\rightarrow$ line\\
\textbf{species $\rightarrow$ RelatedTo $\rightarrow$ race}\\
mines $\rightarrow$ FormOf $\rightarrow$ mine $\rightarrow$ AtLocation $\rightarrow$ home\\
species $\rightarrow$ RelatedTo $\rightarrow$ kingdom $\rightarrow$ RelatedTo $\rightarrow$ queen $\rightarrow$ RelatedTo $\rightarrow$ person\\
species $\rightarrow$ RelatedTo $\rightarrow$ kingdom $\rightarrow$ DerivedFrom $\rightarrow$ king $\rightarrow$ RelatedTo $\rightarrow$ master $\rightarrow$ RelatedTo \\ $\rightarrow$ young}\\
\hline
  \end{tabular}
 \caption{Example 2 selected commonsense paths.}
 \label{tab:extractedex2}
 \end{small}
\end{table*}
\begin{figure*}[h]
  \centering
  \includegraphics[width=0.9\textwidth,trim={0 2cm 0 0},clip]{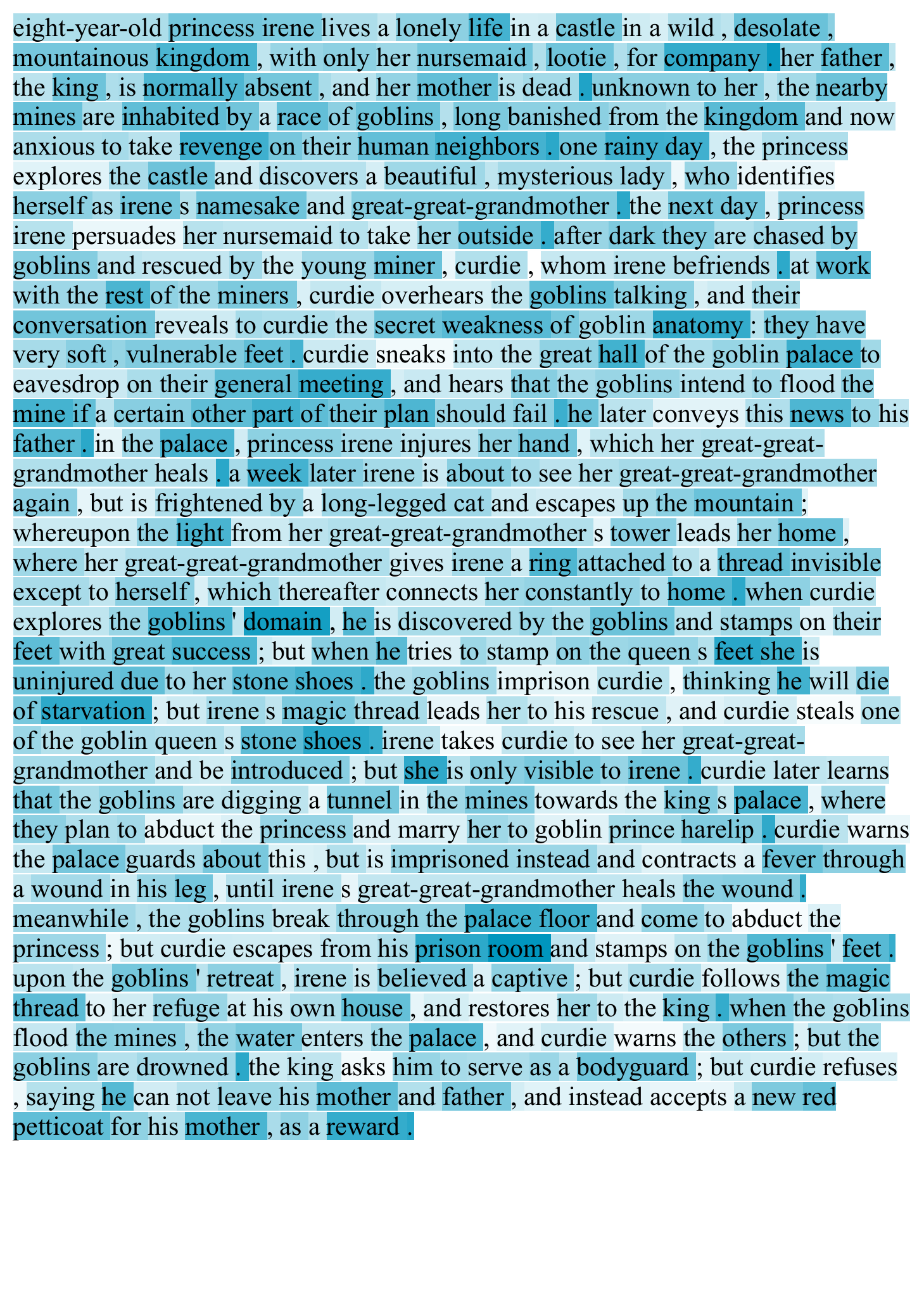}
 \caption{Example 2 visualized activation values of first attention hop ($1 - {\bf z}_1$).}
\end{figure*}
\begin{figure*}[h]
  \centering
  \includegraphics[width=0.9\textwidth,trim={0 2cm 0 0},clip]{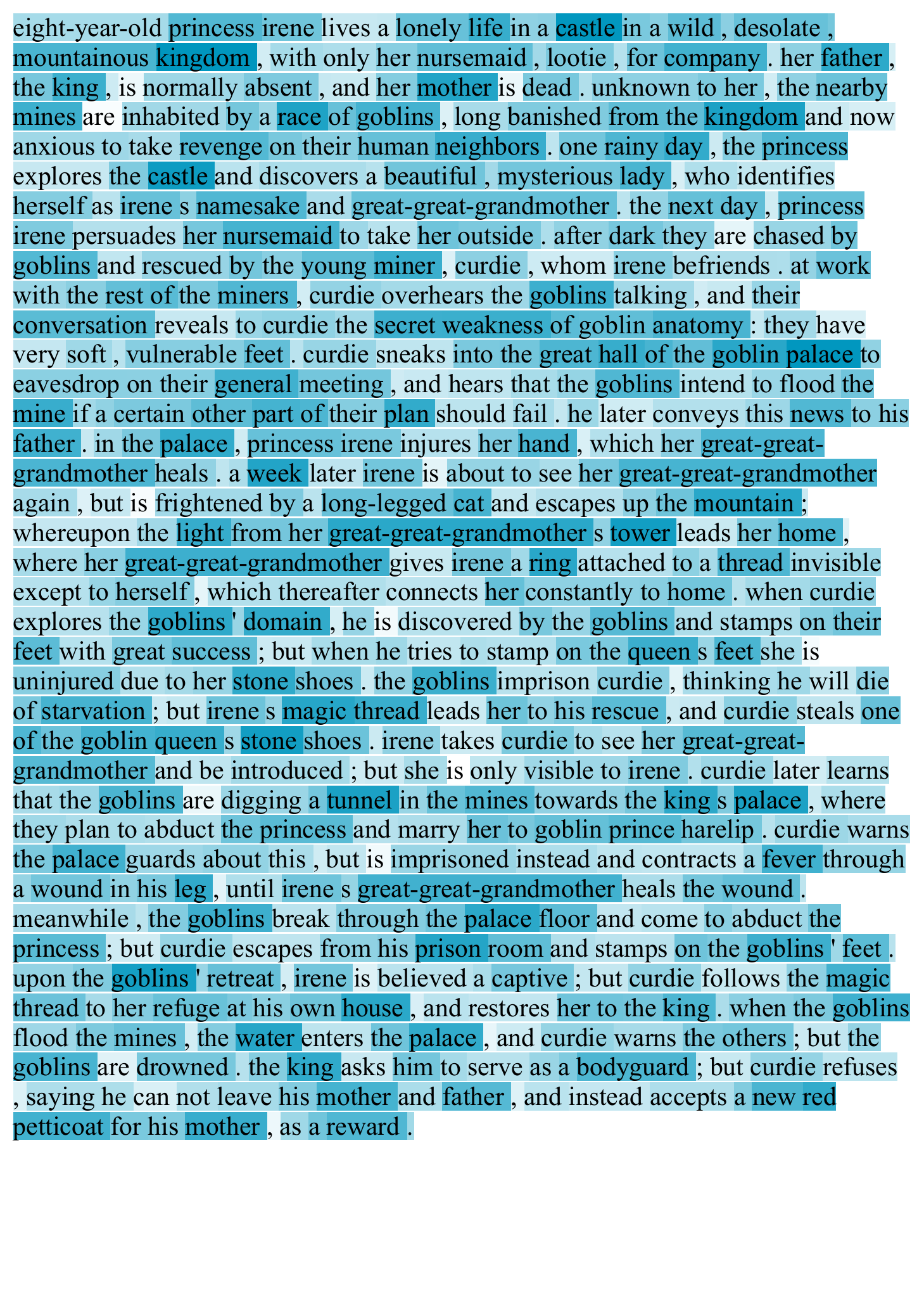}
 \caption{Example 2 visualized activation values of second attention hop ($1 - {\bf z}_2$).}

\end{figure*}
\begin{figure*}[h]
  \centering
  \includegraphics[width=0.9\textwidth,trim={0 2cm 0 0},clip]{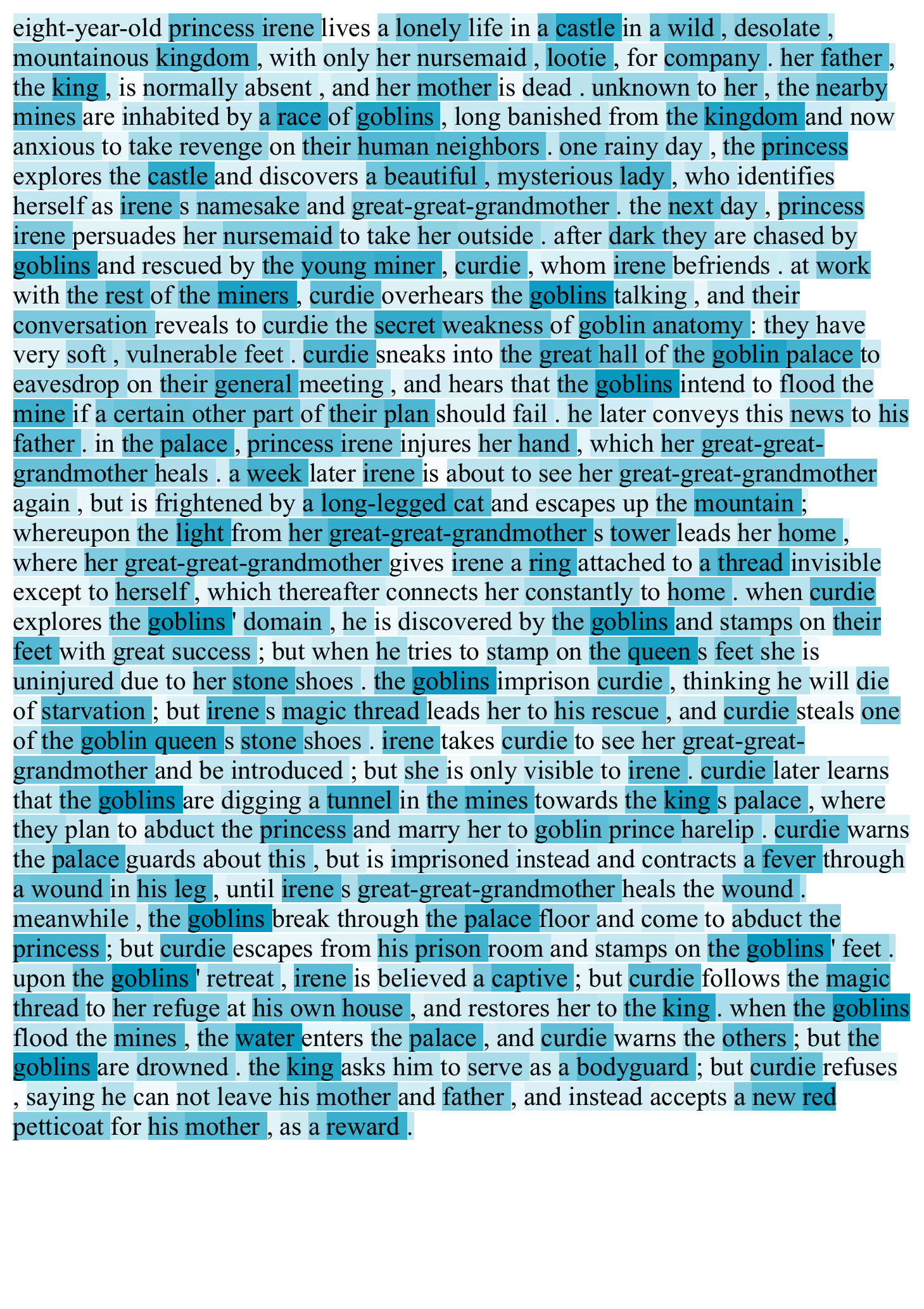}
 \caption{Example 2 visualized activation values of third attention hop ($1 - {\bf z}_3$).}
\end{figure*}
\begin{table*}[!h]
	\centering
    \begin{small}
  \begin{tabular}{|p{0.15\textwidth}|p{0.8\textwidth}|} 
    \hline
    Question & \makecell[l]{What \textbf{duty} does ruth have to fulfill when her aunt dies?\\} \\
    Context &  \makecell[l]{\\"..ruth anvoy, a young american woman with a wealthy father, comes to britain to \\ visit her widowed aunt lady coxon.."\\
"..having made a promise to her now-deceased husband, \textbf{lady} coxon has for \\ years been seeking to bestow a sum of 13,000 pounds upon a talented \\ intellectual whose potential has been hampered by lack of \textbf{money}. having failed to \\ find such a \textbf{person}, lady coxon tells anvoy that upon her death the \textbf{money} will be \\ left to her, and she \textbf{must} carry on the  quest.."\\
"..anvoy, having lost nearly all her wealth, has only the 13,000 pounds from \\ lady coxon, with a \textbf{moral} but not legal \textbf{obligation} to give it away.."\\
"..she awards the coxon fund to saltram, who lives off it exactly as he lived off \\ his friends, producing nothing of intellectual value.."\\ \\}  \\

     Answers    & \makecell[l]{she must give away the 13,000 pounds to an appropriate recipient. /  \\ bestow 13000 to the appropriate \textbf{person}} \\
    \makecell[l]{Extracted \\ Commonsense} & \makecell[l]{\\
duty $\rightarrow$ RelatedTo $\rightarrow$ moral $\rightarrow$ Antonym $\rightarrow$ immoral\\
\textbf{duty $\rightarrow$ RelatedTo $\rightarrow$ time $\rightarrow$ IsA $\rightarrow$ money}\\
duty $\rightarrow$ RelatedTo $\rightarrow$ time $\rightarrow$ IsA $\rightarrow$ money $\rightarrow$ AtLocation $\rightarrow$ church\\
duty $\rightarrow$ DistinctFrom $\rightarrow$ off\\
\textbf{duty $\rightarrow$ RelatedTo $\rightarrow$ time $\rightarrow$ IsA $\rightarrow$ money $\rightarrow$ CapableOf $\rightarrow$ pay $\rightarrow$ bills} \textbf{$\rightarrow$ MotivatedByGoal} \\ \textbf{$\rightarrow$ must}\\
duty $\rightarrow$ RelatedTo $\rightarrow$ time $\rightarrow$ IsA $\rightarrow$ money $\rightarrow$ AtLocation $\rightarrow$ church $\rightarrow$ RelatedTo $\rightarrow$ house\\
duty $\rightarrow$ RelatedTo $\rightarrow$ must $\rightarrow$ RelatedTo $\rightarrow$ having $\rightarrow$ RelatedTo $\rightarrow$ estate $\rightarrow$ RelatedTo $\rightarrow$ real\\
her $\rightarrow$ RelatedTo $\rightarrow$ woman $\rightarrow$ RelatedTo $\rightarrow$ lady $\rightarrow$ RelatedTo $\rightarrow$ plate $\rightarrow$ Antonym $\rightarrow$ her\\
duty $\rightarrow$ RelatedTo $\rightarrow$ moral $\rightarrow$ RelatedTo $\rightarrow$ will $\rightarrow$ RelatedTo $\rightarrow$ choose $\rightarrow$ IsA $\rightarrow$ decide\\
duty $\rightarrow$ RelatedTo $\rightarrow$ must $\rightarrow$ RelatedTo $\rightarrow$ having $\rightarrow$ RelatedTo $\rightarrow$ estate\\
\textbf{duty $\rightarrow$ RelatedTo $\rightarrow$ obligation }\\
duty $\rightarrow$ RelatedTo $\rightarrow$ moral $\rightarrow$ RelatedTo $\rightarrow$ will $\rightarrow$ IsA $\rightarrow$ purpose\\
her $\rightarrow$ RelatedTo $\rightarrow$ but $\rightarrow$ DistinctFrom $\rightarrow$ only $\rightarrow$ RelatedTo $\rightarrow$ child $\rightarrow$ RelatedTo $\rightarrow$ particularly\\
her $\rightarrow$ RelatedTo $\rightarrow$ person $\rightarrow$ RelatedTo $\rightarrow$ others $\rightarrow$ RelatedTo $\rightarrow$ people\\
her $\rightarrow$ Antonym $\rightarrow$ him $\rightarrow$ RelatedTo $\rightarrow$ he $\rightarrow$ RelatedTo $\rightarrow$ person $\rightarrow$ Desires $\rightarrow$ conversation\\
\textbf{her $\rightarrow$ RelatedTo $\rightarrow$ woman $\rightarrow$ RelatedTo $\rightarrow$ lady}\\
her $\rightarrow$ RelatedTo $\rightarrow$ woman $\rightarrow$ RelatedTo $\rightarrow$ she\\
duty $\rightarrow$ RelatedTo $\rightarrow$ must $\rightarrow$ RelatedTo $\rightarrow$ having $\rightarrow$ RelatedTo $\rightarrow$ own $\rightarrow$ RelatedTo $\rightarrow$ having\\
\textbf{her $\rightarrow$ RelatedTo $\rightarrow$ person $\rightarrow$ DistinctFrom $\rightarrow$ man $\rightarrow$ Antonym $\rightarrow$ people}\\
her $\rightarrow$ RelatedTo $\rightarrow$ but $\rightarrow$ DistinctFrom $\rightarrow$ only $\rightarrow$ RelatedTo $\rightarrow$ child\\
\textbf{her $\rightarrow$ Antonym $\rightarrow$ him $\rightarrow$ RelatedTo $\rightarrow$ he $\rightarrow$ RelatedTo $\rightarrow$ person}\\
\textbf{her $\rightarrow$ Antonym $\rightarrow$ his $\rightarrow$ RelatedTo $\rightarrow$ him $\rightarrow$ RelatedTo $\rightarrow$ person}\\
    } \\
    \hline
  \end{tabular}
 \caption{Example 3 selected commonsense paths.}
 \label{tab:extractedex3}
 \end{small}
\end{table*}
\begin{figure*}[h]
  \centering
  \includegraphics[width=0.9\textwidth,trim={0 4cm 0 0},clip]{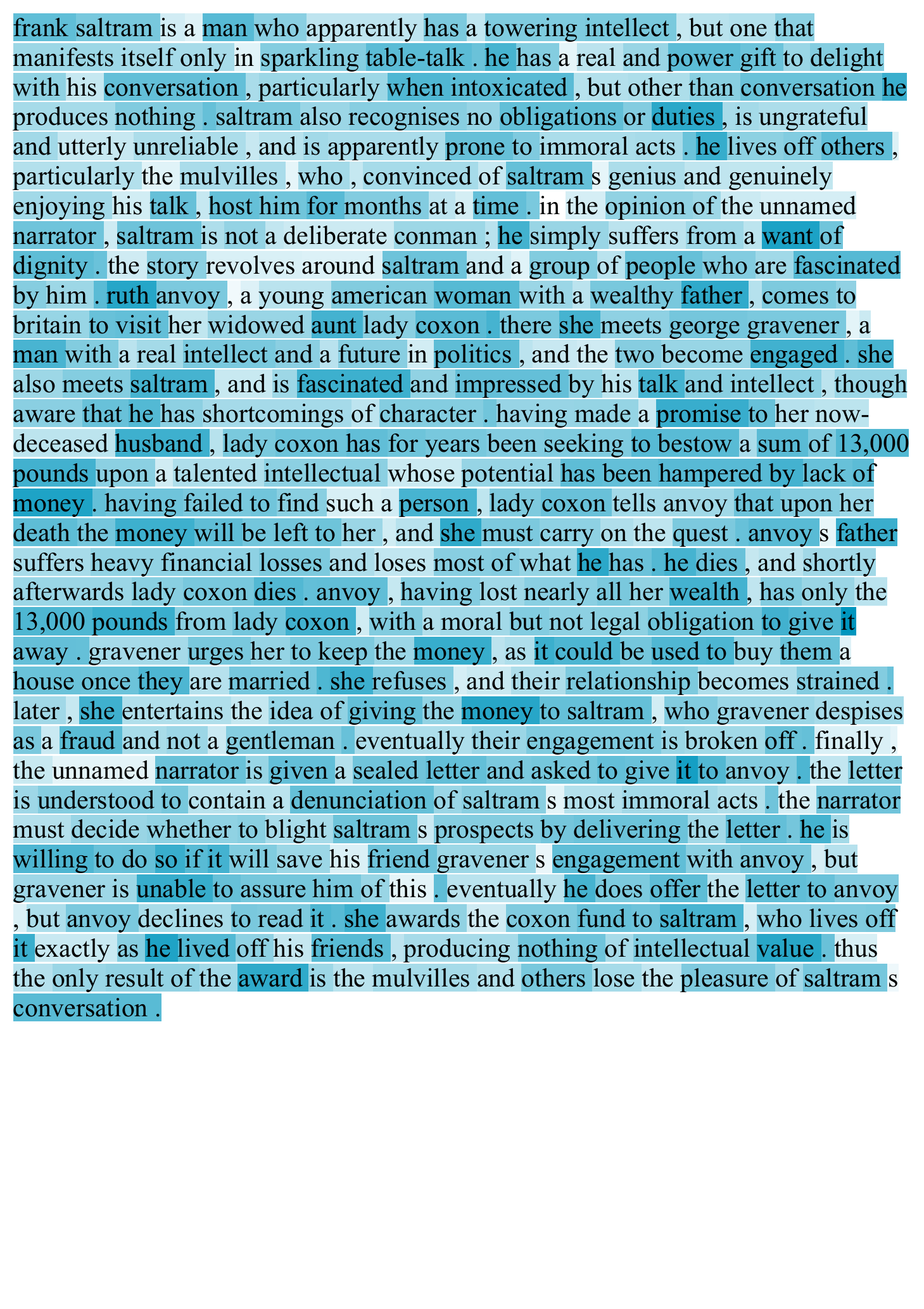}
 \caption{Example 3 visualized activation values of first attention hop ($1 - {\bf z}_1$).}
\end{figure*}
\begin{figure*}[h]
  \centering
  \includegraphics[width=0.9\textwidth,trim={0 4cm 0 0},clip]{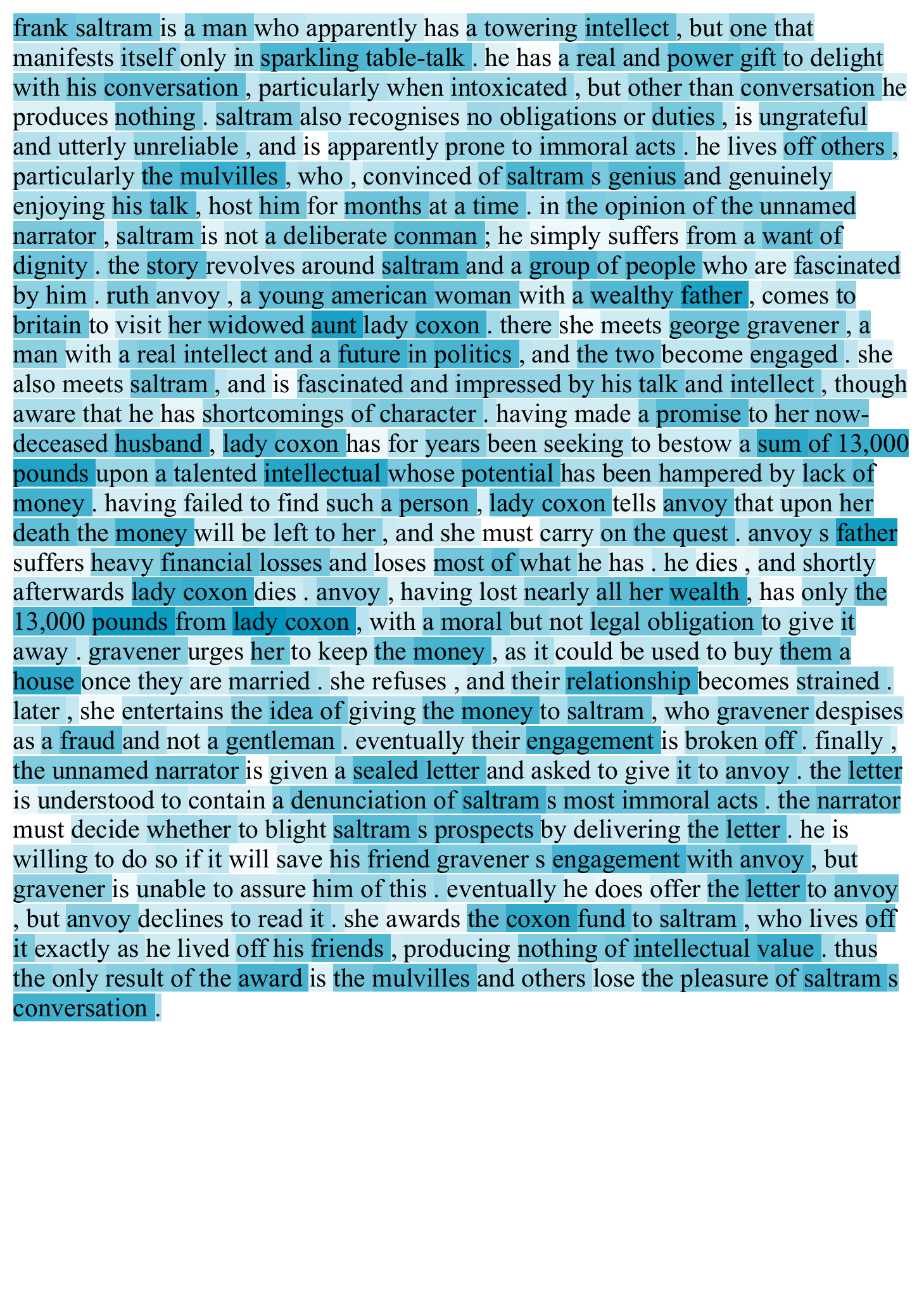}
 \caption{Example 3 visualized activation values of second attention hop ($1 - {\bf z}_2$).}
\end{figure*}
\begin{figure*}[h]
  \centering
  \includegraphics[width=0.9\textwidth,trim={0 4cm 0 0},clip]{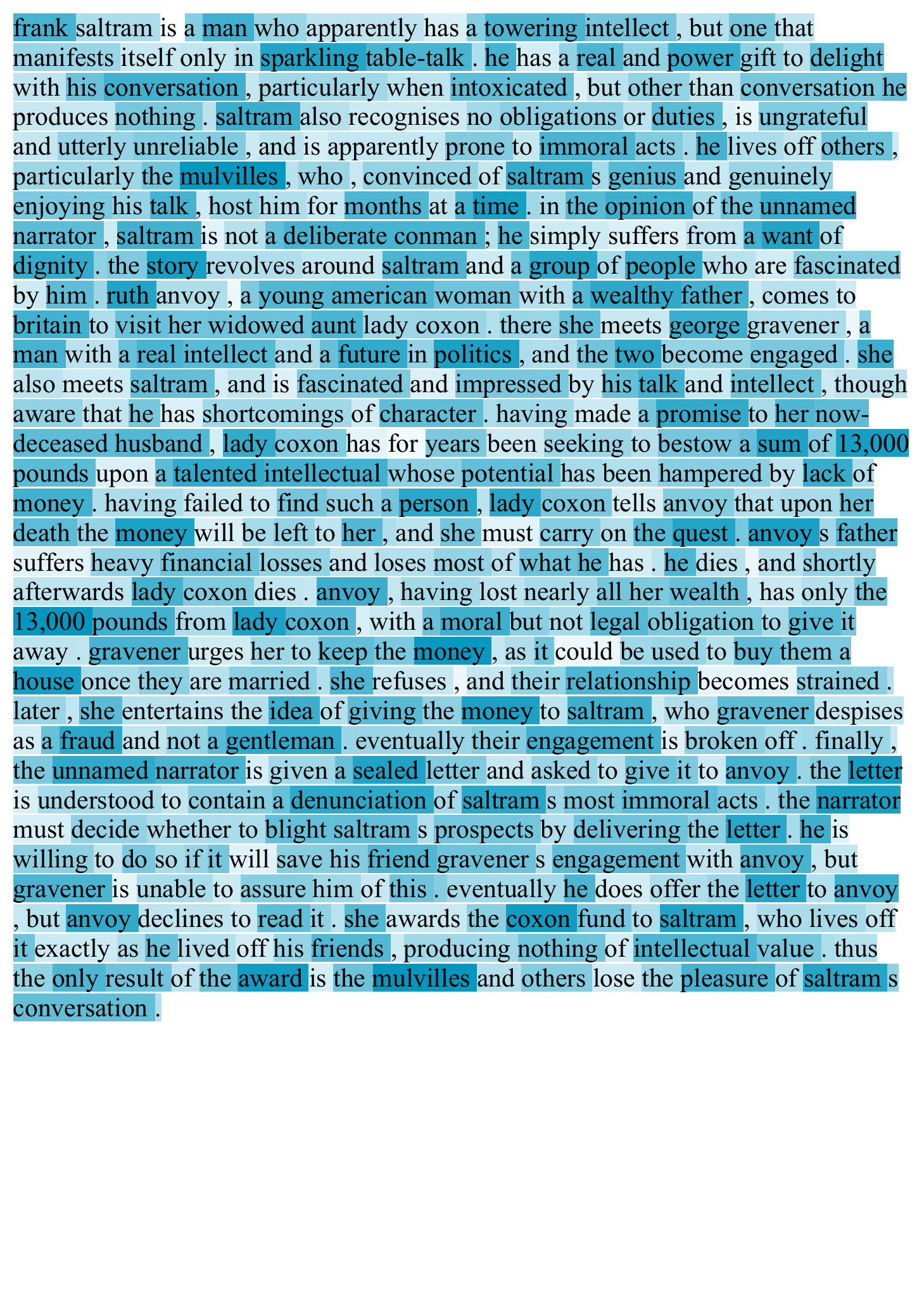}
 \caption{Example 3 visualized activation values of third attention hop ($1 - {\bf z}_3$).}
\end{figure*}

\subsection{Commonsense Extraction Examples}
In Tables~\ref{tab:extractedex1},~\ref{tab:extractedex2}, and~\ref{tab:extractedex3} (see next page), we demonstrate extracted commonsense examples for questions that require commonsense to reach an answer. We bold words in the question and in the extracted commonsense in cases where the commonsense knowledge explicitly bridges gaps between implicitly connected words in the context or question. The relevant context is also displayed, with context words that are key to answering the question (via commonsense) marked in bold. These are then followed by a context visualization described in the next section.

\subsection{Commonsense Integration Visualization}
We also visualize how much commonsense information is integrated into each part of the context by providing a visualization of the ${\bf z}_i$ value (see end of~\secref{sec:commonsense_model_inc} of main file) for $i \in \{1,2,3\}$, which is the gate value signifying how much commonsense-attention representation is used in the output context representation. In the following examples (next page), we use shades of blue to represent the average of $(1 - {\bf z}_i)$ at each word in the context (normalized within each hop), with deeper blue indicating the use of more commonsense information. As a general trend, we see that in the earlier hops, words which are near tokens that occur in both the context and commonsense paths have high activation, but the activation becomes more focused on the passage's key words w.r.t. the question, as the number of hops increase.

\end{document}